\documentclass[preprint,review]{elsarticle}

\usepackage[export]{adjustbox}

\usepackage{hyperref}

\usepackage{siunitx}

\journal{Knowledge-Based Systems}

\begin{document}

\begin{frontmatter}

\title{Optical images-based edge detection in \\ Synthetic Aperture Radar images}


\author[mymainaddress]{Gilberto P. Silva Junior}

\author[mysecondaryaddress]{Alejandro C. Frery}

\author[mymainaddress]{Sandra Sandri\corref{mycorrespondingauthor}}
\cortext[mycorrespondingauthor]{Corresponding author}
\ead{sandra.sandri@inpe.br}

\author[mytertiaryaddress]{Humberto Bustince}
\author[mytertiaryaddress]{Edurne Barrenechea}
\author[mytertiaryaddress]{C\'edric Marco-Detchart}

\address[mymainaddress]{Laborat\'orio Associado de Computa\c{c}\~ao e Matem\'atica Aplicada, 
Instituto Nacional de Pesquisas Espaciais (LAC/INPE), Av. dos Astronautas, 1758, 
12227--010,  \\
S\~ao Jos\'e  dos Campos, SP -- Brazil}
\address[mysecondaryaddress]{Laborat\'{o}rio de Computa\c{c}\~{a}o Cient\'{i}fica e An\'{a}lise Num\'{e}rica (LaCCAN/UFAL), Universidade Federal de Alagoas, Av.\ Lourival Melo Mota, s/n,
57072-970, \\ 
Macei\'{o}, AL -- Brazil}
\address[mytertiaryaddress]{Departamento de Autom\'atica y Computaci\'on, Universidad P\'ublica de Navarra, Campus Arrosad\'{\i}a,
31006, Pamplona, Spain}

\begin{abstract}
We address the issue of adapting optical images-based edge detection techniques for use in Polarimetric Synthetic Aperture Radar (PolSAR) imagery. 
We modify the gravitational edge detection technique (inspired by the Law of Universal Gravity) proposed by Lopez-Molina et al, using the  non-standard neighbourhood configuration proposed by Fu et al, to reduce the speckle noise in polarimetric SAR imagery.
We compare the modified and unmodified versions of the gravitational edge detection technique with the well-established one proposed by Canny, as well as with a recent multiscale fuzzy-based technique proposed by Lopez-Molina et al.
We also address the issues of aggregation of gray level images before and after edge detection and of filtering.
All techniques addressed here are applied to a mosaic built using class distributions obtained from a real scene, 
as well as to the true PolSAR image; the mosaic results are assessed using Baddeley's Delta Metric.
Our experiments show that modifying the gravitational edge detection technique with a non-standard neighbourhood configuration  produces better results than the original technique, as well as the other techniques used for comparison. 
The experiments show that adapting edge detection methods from Computational Intelligence for use in PolSAR imagery is a new field worthy of exploration. 
\end{abstract}

\begin{keyword}
Edge detection \sep SAR images\sep Computational Intelligence\sep Gravitational method 
\end{keyword}

\end{frontmatter}

\section{Introduction}\label{sec:intro}

Edge detection seeks to  identify sharp differences automatically in the information associated with adjacent pixels in an image \cite{Gonzalez2006}. 
Edge detection for optical images is nowadays quite an established field. 
It is traditionally carried out using gradient-based techniques, 
such as the well-known Canny algorithm~\cite{CANNY1986}.
Techniques based on Computational Intelligence have also been proposed in the recent literature. 
Sun et al \cite{Sun2007} proposed the gravitational edge detection method, inspired by Newton's Universal Law of Gravity. 
Lopez-Molina et al \cite{LopezMolina2010} proposed a fuzzy extension for this technique, allowing the use of T-norms, a large class of fuzzy operators; they also proposed small modifications in the basic formalism (see Section~\ref{sec:relwork}).
D\u{a}nkov\'a  et al \cite{Dankova2011} proposed the use of a fuzzy-based function, the F-transform;  the original universe of functions is transformed into a universe of their “skeleton models” (vectors of F-transform components), making further computations easier to perform.
Barrenechea et al \cite{barrechea11} proposed the use of interval-valued fuzzy relations for edge detection, using a T-norm and a T-conorm to produce a fuzzy edge image, that is then binarized. 
This approach was extended by Chang and Chang \cite{chang-chang13}. First of all, two new images are created---one rather dark and the other rather bright---by applying two different parameters on the linear combinations of the images obtained using $\min$ and $\max$ operators, respectively.
Then, the fuzzy edge image is created by the difference between these two new images.  
Another recent approach from Computational Intelligence is the multiscale edge detection method proposed by Lopez-Molina et al 
 \cite{lopez-molina-13-multiscale}, using Sobel operators for edge
extraction and the concept of Gaussian scale-space. 
  
SAR sensors are not as adversely affected by atmospheric conditions and the presence of clouds as optical sensors. 
Moreover, unlike the optical counterparts, SAR sensors can be used at any time of day or night. 
For these reasons,  remote sensing applications using  SAR imagery have been growing over the years~\cite{Mott2006}.
SAR images, however, contain a great amount of noise, known as \textit{speckle}, that degrades the visual quality of the images. 
Caused by inherent characteristics of  radar technology, this multiplicative non-Gaussian noise is proportional to the intensity of the received signal.

Contrary to what happens with optical images, there are still few algorithms specifically dedicated to SAR images \cite{Fu2012}.
One interesting means to create edge detection algorithms for SAR images is to modify  those created for optical images.
However, the use of these methods on SAR images is not straightforward, due to speckle. 
One can either adapt optical image techniques to meet SAR data properties, or first preprocess the images using filters and then apply the original optical techniques. 

The main purpose of our study is to investigate the application of the gravitational edge detection,
Here we modify the original 3$\times$3 window:
the value in each cell in the window is no longer the original one, 
but the aggregation of a set of neighbouring pixels, according to the larger $9\times9$ neighbourhood configuration proposed by Fu et al~\cite{Fu2012}.
We propose a typology of experiments to study the behaviour of the modified edge detection method, considering  polarization,  image aggregation, and image binarization. 
We focus on the use of the following processes:
DAB (edge Detection on non-binary images, Aggregation of the resulting non-binary  images, Binarization) 
and
ADB (Aggregation of non-binary  images, edge Detection on the resulting non-binary  image, Binarization). 

We also investigate the use of noise-reduction filters in preprocessing the images, by making use of the the well-known Enhanced Lee filter~\cite{Lopes90} and a filter recently proposed by Torres et al~\cite{TorresPolarimetricFilterPatternRecognition}. 

Barreto et al~\cite{Barrreto2013} describe a classification experiment, based on  a full polarimetric image from an agricultural area in the Amazon region in Brazil.
In that study,  the authors estimated the parameters for probability distributions associated to each of the classes of interest, such as  water and different types of vegetation and their phenology.  
They assessed their results in an image formed by a mosaic of the classes, with pixel values generated using the parameters found for each class.
We apply all techniques addressed in this study on twenty simulated mosaics, using the parameters estimated in~\cite{Barrreto2013}, considering amplitude images derived from different polarizations. 
We assess the quality of the results, according to Baddeley's Delta Metric (BDM)~\cite{Baddeley1992}. 

We also apply the methods on the real images, but assessment is only visual.
We compare our results with those produced by the use of Canny's algorithm~\cite{CANNY1986} and the recently proposed multiscale method by Lopez-Molina et al~\cite{lopez-molina-13-multiscale}.
 
The present study is an extended version of~\cite{SilvaFLINS14}, in which some of the main ideas of this paper were first delineated.
However, the present study and~\cite{SilvaFLINS14} differ in the scope of the proposed approach as well as in the reliability of the results. 
Indeed, in~\cite{SilvaFLINS14}, only one simulated image was used in the experiments and only Canny's technique was compared to its results. 
Moreover, in the previous paper we only addressed the edge detection of the image resulting from the aggregation of  the three simulated polarization  images. 
In our first paper only ADB was addressed; 
edge detection on the individual polarization images as well as DAB strategy were not considered. 

The results from our current study show that adapting edge detection methods from Computational Intelligence to use in radar imagery is a new field worthy of exploration. 
In particular, our experiments show that modifying the gravitational method with Fu's $9 \times 9$ neighbourhood produces better results than the unmodified method. 
They also show the importance of filtering when adapting edge detection techniques from optical to radar images. 

\section{Basic concepts on SAR images}

Optical and SAR sensors measure the amount of energy reflected by a target in various bands of the electromagnetic spectrum. 
The bands  employed in most  imaging radars use frequencies in the \SI{2}{\mega\hertz} to \SI{12.5}{\giga\hertz} range, with wavelengths ranging from \SI{2.4}{\centi\meter} to \SI{1}{\meter}. 
In this study, we used only the 
L-band with wavelengths of [\SI{30}{\centi\meter}, \SI{1}{\meter}] and frequencies of [\SI{1}{\mega\hertz}, \SI{2}{\giga\hertz}].

\begin{figure}[hbt]
	\centering
	\includegraphics[scale=.2,frame]{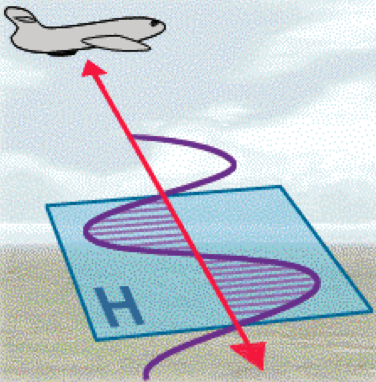} $\;\;$
	\includegraphics[scale=.2,frame]{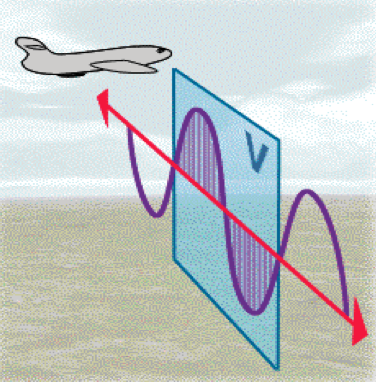}
	\caption{Horizontal and vertical signal polarizations transmitted by an antenna. Source: \cite{Meneses2012}}
	\label{fig:pol}
\end{figure}

SAR systems generate the image of a target area by moving along a usually linear trajectory, and transmitting pulses in lateral looks towards the ground, in either horizontal (H) or vertical (V) polarizations~\cite {Richards2009}, respectively denoted as H and V  (see Figure~\ref{fig:pol}). 
In the past, the reception of the transmitted energy was made ​​solely on the same polarization of the transmission, generating images in the HH and VV polarizations. 
Currently, with the advent of polarized and fully polarimetric radars (PolSAR - \emph{Polarimetric Synthetic Aperture Radar}), information about intensity and phase of the cross signals are also obtained, generating images relating to HV and VH polarizations. 
Usually, applications only consider the HH, VV, and HV polarizations.

The imaging can be obtained by gathering all the intensity and phase information data 
from the electromagnetic signal after it has been backscattered by the target in a given polarization \cite{LeePottier2009}. 
Each polarization in a given a scene generates a complex image, which can be thought of as two images,  containing the real and imaginary values for the pixels, respectively. 

\begin{figure}[hbt]
	\centering
	\includegraphics[width=.3\linewidth,frame]{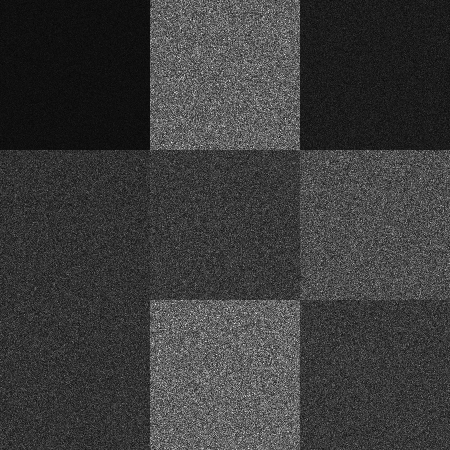}
	\includegraphics[width=.3\linewidth,frame]{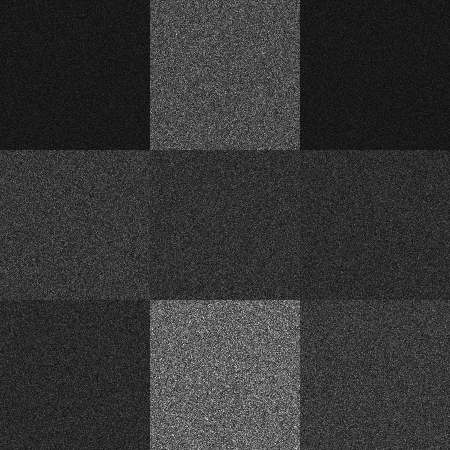}
	\includegraphics[width=.3\linewidth,frame]{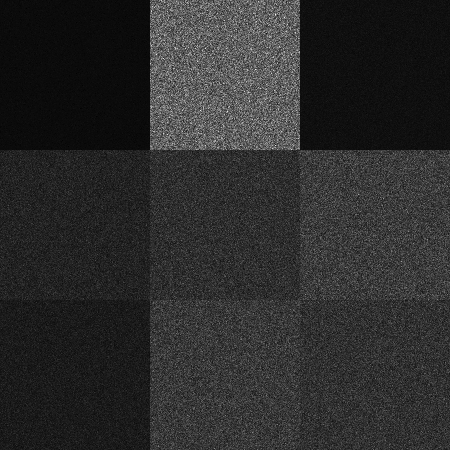}\hfill
	$A_{HH} \;\;\;\;\;\;\;\;\;\;\;\;\;\;\;\;\;\;\;\;\;\;\;\;\;\; A_{HV} \;\;\;\;\;\;\;\;\;\;\;\;\;\;\;\;\;\;\;\;\;\;\;\;\;\; A_{VV}$
	\caption{Amplitude images for polarizations HH, VV, and HV from the same scene}
	\label{fig:int-amp}
\end{figure}

We denote the complex images from HH, VV, and HV polarizations  as $S_{HH}$, $S_ {HV}$, and $S_ {VV}$.
Multiplying the vector $[S_ {HH}\; S_ {HV}\; S_ {VV}]$ by its transposed conjugated vector $[S_ {HH}^*\; S_ {HV}^*\; S_ {VV}^*]^t$, 
we obtain a $3 \times 3$ covariance matrix. 
The main diagonal contains intensity values; taking their square root, we obtain amplitude values. 
We denote the intensity images by $I_{HH}$, $I_{HV}$, and $I_{VV}$ and their corresponding amplitude counterparts by $A_{HH}$, $A_{HV}$, and $A_{VV}$. 
In this paper, we  only considered the amplitude images, such as those depicted in Figure~\ref{fig:int-amp}.

Speckle noise is multiplicative, non-Gaussian, and is proportional to the intensity of the received signal.
Speckle degrades the visual quality of the displayed image by sudden variations in image intensity with a salt and pepper pattern, 
as can be seen in Figure~\ref{fig:int-amp}.
It can be reduced with multiple looks in the generation of the complex images, causing degradation in spatial resolution. Another way to reduce noise is to employ filters, as will be discussed in the next section.

In SAR image classification, one often uses samples from the classes in order to estimate the parameters of the distribution believed to underlie each class. 
Synthetic images can then be created using Monte Carlo simulation by taking the realization of the random variable associated to the class of each classified pixel.
This artifice is useful to choose the most apt classifier for a given application: instead of relying solely on the original image, one takes the classifier that obtains the best average accuracy on the set of synthetic images. 
This methodology can also be used in other tasks, such as edge detection.

\section{Related Work}\label{sec:relwork}

One of the most successful edge detection algorithms for optical images was proposed by Canny~\cite{CANNY1986}, based on the following guidelines: 
i)~the algorithm should mark as many real edges in the image as possible; 
ii)~the marked edges  should be as close as possible to the edge in the real image; 
iii)~a given edge in the image should only be marked  once; and 
iv)~image noise should not create false edges.
It makes use of numerical optimization to derive optimal operators for ridge and roof edges. 
The usual implementation of this method uses a $3 \times 3$ neighbourhood. 

A more recent multi-scale edge detection method  was proposed by Lopez-Molina et al~\cite{lopez-molina-13-multiscale}, 
using  Sobel operators for edge extraction and the concept of Gaussian scale-space. 
More specifically, the Sobel edge detection method is applied on increasingly
smoother versions of the image. 
Then, the edges which appear on different scales are combined by performing coarse-to-fine
edge tracking.

The gravitational edge detection approach was first proposed by Sun et al~\cite{Sun2007} and applied to optical images. It is based on Newton's Universal Law of Gravity, described by Equation~(\ref{eq:newton}):
\begin{equation}
f_{1,2} = G \times \frac{m_1 \times m_2}{\|\vec{r}\|^2}  \times  \frac{\vec{r}}{\|\vec{r}\|},
\label{eq:newton}
\end{equation}
where $m_1$ and $m_2$ are the masses of two bodies; 
$\vec{r}$ is the vector connecting them; 
$\vec{f}_{1,2}$ is the gravitational force between them; 
$\|.\|$  denotes the magnitude of a vector; 
and $G$ is the gravitational constant. 
In the analogy proposed by Sun et al~\cite{Sun2007}; 
the bodies are the gray level values of pixels in a grid;
$G$ is a function of the values of the pixels in a given window; 
the distance between any two adjacent pixels is equal to 1; 
and, when computing the resulting force of the pixel in the center of a window; 
the pixels outside that window are considered negligible. 
Lopez-Molina et al~\cite{LopezMolina2010} extended this technique, 
proposing the use of a Triangular Norm~\cite{DP88} in place of the product between the two masses\footnote{Triangular norm operators are mappings from $[0, 1]^2$ to $[0, 1]$, that are commutative, associative, monotonic, and have $1$ as neutral element.}, 
by first normalizing the gray level values to $[0,1]$. 
The authors treat edges as fuzzy sets for which membership degrees are extracted from the resulting gravitational force on each pixel. 
They take $G$ as a normalization constant, calculated so as to guarantee that the resulting forces   lie in [0,1]. 
Also, in the normalization of gray level values into [0,1], a small value $\delta q$ is added beforehand to both the numerator and denominator  so as avoid pixels with value 0, 
which would have too strong an effect on neighbouring pixels. 
The authors used  $3 \times 3$ and $5 \times 5$ windows as well as several prototypical triangular norms. 

The so-called Lee  (or sigma) filter introduced in 1983~\cite{LeeFilter1983}, is still in use today due to its simplicity, 
its effectiveness in speckle reduction, and its computational efficiency. 
It is based on the fact that, under Gaussian distribution, approximately $95.5\%$ of the probability is concentrated within two standard deviations from the mean.
The filter estimates the mean and the standard deviation of samples around each pixel, and only those values within this interval are used to compute the local mean. 
Lopes et al~\cite{Lopes90} proposed an adaptive  version for this filter, here referred to as ``Enhanced Lee''.

Torres et al~\cite{TorresPolarimetricFilterPatternRecognition} recently proposed a nonlocal means approach for PolSAR image speckle reduction based on stochastic distances; the method can be tailored to any distribution, both univariate of multi-variate. 
It consists of comparing the distributions which describe the central observation for each pixel, and each of the observations which comprise a search region. 
The comparison is made through a goodness-of-fit test, and the $p$-value of the test statistic is used to define the convolution matrix which will define the filter: the higher the $p$-value the larger the confidence and, thus, the importance, each observation will have in the convolution. 
In Torres et al's proposal, the tests are derived from
$h$-$\phi$ divergences between multi-look scaled complex Wishart distributions for fully PolSAR data~\cite{Frery2014}.
Their results are competitive with classical and advanced polarimetric filters, with respect to  usual quantitative measures of quality. 

\begin{figure}[htb]
\centering
\includegraphics[scale=.25,frame]{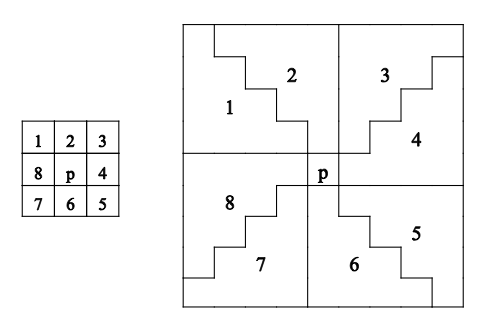}
\caption{Standard $3 \times 3$ and Fu's neighbourhood \cite{Fu2012}} 
\label{fig:vizinhaca_Fu}
\end{figure}

Fu et al~\cite{Fu2012} proposed a statistical edge detector suitable for SAR images  which uses the squared successive difference of averages to estimate the edge strength from the sliding window. 
An interesting feature of this paper is the proposal of a specific type of $9 \times 9$ neighbourhood, shown in Figure~\ref{fig:vizinhaca_Fu}.
In a previous paper~\cite{SilvaFLINS14}, we proposed a modification of the gravitational approach using Fu et al's neighbourhood: given a central pixel in a $3 \times 3$ window in an image, the values considered for the surrounding pixels in the window are no longer the ones in the original image, but the mean values in this new configuration.

\section{Materials and Methods}\label{sec:metod}

We compare the edge detection methods proposed by Canny~\cite{CANNY1986} and by Lopez-Molina et al~\cite{lopez-molina-13-multiscale} to the modified gravitational approach using the product T-norm,  followed by thresholding. 
The effect of preprocessing the images through filtering is also studied, using the filter described by Torres et al~\cite{TorresPolarimetricFilterPatternRecognition} and the Enhanced Lee filter~\cite{Lopes90}.
We study the behaviour of Lopez-Molina's method with the usual $3 \times 3$ window as well as a modified version of this approach, proposed in a previous paper~\cite{SilvaFLINS14}, involving the neighbourhood proposed by Fu et al~\cite{Fu2012}.

The input for Canny's and Lopez-Molina's edge detector are images in, respectively, $\{0, \dots, 255\}$ and $[0,1]$.
Image values are, thus, mapped into these sets prior to edge detection.
For the Lopez-Molina methods (the original and modified versions), we normalize further to $[0,1]$, using  $\delta q=1$ and making $q'= ({q + 1})({255+1})^{-1}$, where $q$ and  $q'$ are the old and new value of a given pixel, respectively.

\subsection{Working image}

\begin{figure}[hbt]
\centering
a) \includegraphics[scale=.50]{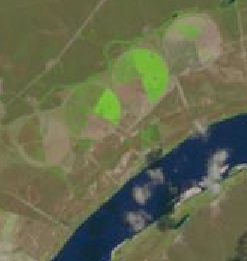}
b) \includegraphics[scale=.35]{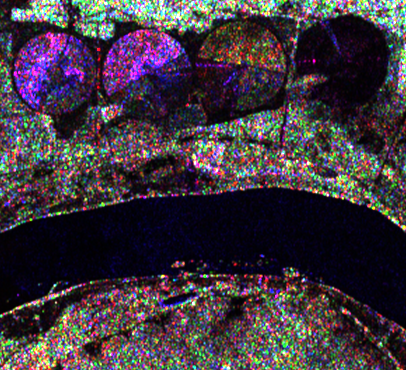}\\
\footnotesize  
\caption{Images derived from a scene in Bebedouro in Brazil (not registered): 
a) Landsat RGB composition and
b) SAR L-band RGB composition
(source: \cite{Barrreto2013})}
 \label{aba:fig_image}
\end{figure}

\begin{figure}[hbt]
\centering
\includegraphics[scale=.19]{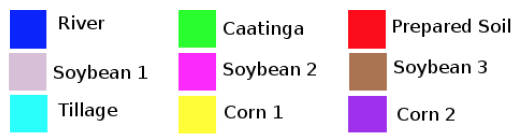}
$\;\;\;\;\;\;\;\;\;\;\;\;\;\;\;\;\;\;\;\;\;$
$\;\;\;\;\;\;\;\;\;\;\;\;\;\;\;\;\;\;\;\;\;$\\ 
a)\includegraphics[scale=.38]{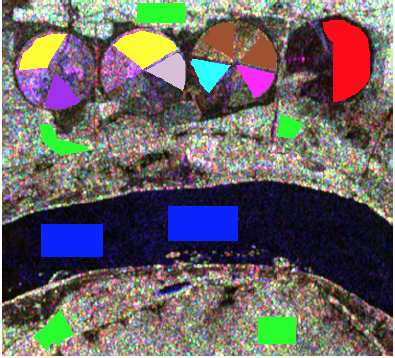}
b)\includegraphics[scale=.30]{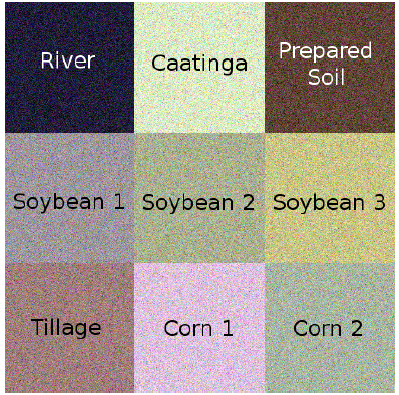} 
\footnotesize  
\caption{Images derived from a scene in Bebedouro in Brazil: 
a) training samples used to generate Wishart distributions 
and 
b) synthetic mosaic images generated using the Wishart distributions estimated in \cite{Barrreto2013} from image samples
(source: \cite{Barrreto2013})}
 \label{aba:fig_samples}
\end{figure}

We apply the methods on data derived from a fully polarimetric image, presented by Barreto et al~\cite{Barrreto2013}, 
from an agricultural area in the Amazon region in Brazil (see Figure \ref{aba:fig_image}).  
The authors describe a classification experiment using classes of interest from that area, such as  water and different types of crops and natural vegetation, at different stages of growth.  
Samples from the classes from band L are used to estimate the parameters of the complex Wishart distribution associated to each class.  
The results are assessed using a mosaic with the classes that was created using the derived Wishart distributions. 
Figure~\ref{aba:fig_samples}  illustrates the approach. 
For our study  we apply the edge detection methods on twenty independently simulated mosaics amplitude images, 
using the parameters estimated in~\cite{Barrreto2013} to  assess the quality of the methods. 

\subsection{Quality assessment}

The quality of the results is assessed by the Baddeley's Delta Metric (BDM) \cite{Baddeley1992}, by comparison with what would be the perfect result, discarding those pixels close to the outer frame.

Let $\bf{x}$ and $\bf{y}$ be two binary images, seen as mappings from $\Lambda$ to $[0,1]$, 
where $\Lambda$ is a set of sites arranged in a grid (positions). 
Let $\rho$ be a metric on $\Lambda$, such as the Euclidean distance, 
and $d(i, A)$ be the distance between a site $i$ and a set  $A \subseteq \Lambda$, defined as 
$$d(i, A)=\min_{j \in A} \rho(i,j).$$ 
Let $b(\bf{x})= \{i \in \Lambda \mid x_i=1\}$ denote the set of foreground sites in $\bf{x}$.
BDM between $\bf{x}$ and $\bf{y}$, denoted as $\Delta_{p,w.}(.,.)$, is then defined as 
\begin{equation}
\Delta_{p,w}({\bf{x}}, {\bf{y}})=
\left(\frac{1}{\mid \Lambda \mid} \sum_{i \in \Lambda} \mid w(d(i,b({\bf x})) - w(d(i,b({\bf y}))\mid^p\right)^{\frac{1}{p}}, 1 \leq p \leq \infty
\label{eq:BDM}
\end{equation}
where $w$ is a  strictly increasing concave function satisfying $w(0)=0$. Here we use $w(t)=t$ and $p=2$, as in~\cite{LopezMolina2010}. 

Throughout the text, we display BDM results in $[0,100]$ instead of $[0,1]$, for the sake of readability.

\section{Proposed methodologies}

Edge detectors use a window around a center pixel to verify whether that pixel belongs to an edge or not. 
When adapting optical image edge detectors to radar imagery, we have to find the means to deal with speckle.  
The main contribution of this study is to modify the original $3\times3$ window used by the edge detection method proposed in~\cite{LopezMolina2010} for use in radar imagery such that
the value in each cell in the window is no longer the original one but the aggregation of set of neighbouring pixels, according to a larger $9\times9$ non-standard neighbourhood proposed by Fu et al~\cite{Fu2012}. 
We here investigate this particular combination of method and neighbourhood, but the same procedure can be applied using other edge detection methods and/or non-standard filters.

Frequently, a single band is used in edge-detection, resulting in a
gray level-image that is then binarized at some point (the usual
implementation of some methods, like Canny's, already involve a
binarization step).
In  radar imagery, very often one deals with more than one band at the same time (e.g. intensity images coming from different polarizations, or complex images in the fully polarimetric case), aiming at using the richness of information to compensate for the speckle noise. 
Therefore, the question of when to aggregate results has to be addressed. 
One may, for instance, first aggregate the bands and then apply the edge detector on the aggregated image, or else apply the edge detector on the individual bands and then aggregate the edge images. These two methods usually yield different results.

Here, we propose a typology for experiments using radar imagery, considering 
different orderings of three steps: edge detection on gray level images, binarization of gray level images, and aggregation of results. 
In the aggregation step, the input may be either gray level or binary images, depending on whether the binarization is made immediately after edge detection or not.  Three strategies can then be envisaged to perform edge detection experiments with radar images: 
\begin{itemize}
\item 
DAB (edge Detection on non-binary images, Aggregation of the resulting non-binary  images, Binarization) and
\item
ADB (Aggregation of non-binary  images, edge Detection on the resulting non-binary  image, Binarization). 
\item
DBA (edge Detection on non-binary images, Binarization, Aggregation of the resulting binary images).
\end{itemize}
Options ABD, BAD and BDA are not considered, since that would mean applying edge detectors on the binary images.

In this work, we focus on the DAB and ADB strategies. For both of them, we use the arithmetic mean to aggregate gray level images. 
Strategy DBA, involving the aggregation of binary images, is left for future study. 

When no aggregation is considered, the strategies are reduced to only edge detection and binarization. 
For example, for the HH, HV and VV polarizations, we obtain strategies DB-HH, DB-HV and DB-HH. 
Note that some methods already incorporate the binarization step in the edge detector. That is for instance the case of all methods discussed previously.
However, to be consistent with the notation, we will denote by ADB the strategy in a method that includes binarization, such as Canny's  and the multi-scale method,  when it is applied to the gray level image resulting from the aggregation of the images from the  HH, HV and VV polarizations.

\section{Experimental Results}\label{sec:expresults}

The output of the Lopez-Molina gravitational method is an image with values in $[0,1]$. 
In order to obtain binary indicators of edges, the authors use a hysteresis transformation. 
Here, we use a simple threshold and search for the value in the $[0.05,0.15]$ interval which produces the best BDM. 
For Canny's method, we search for the best value for  the noise standard deviation parameter $\sigma$ in the interval $[0.3, 1.5]$.
The intervals  above for both Canny and Lopez-Molina are the ones that presented the best results by trial-and-error.
The following parameters were used for the Lopez-Molina multi-scale method, as suggested in \cite{lopez-molina-13-multiscale}:
$\delta_\sigma = 0.25$, $\sigma \in$ 
$\{0.50, 0.75, 1.00,$ 
$1.25, 1.50, 1.75, 2.00,$ 
$2.25, 2.50, 2.75, 3.00, $ 
$3.25, 3.50, 3.75, 4.00\}$. 

We applied two filters (Torres~\cite{TorresPolarimetricFilterPatternRecognition} and Enhanced Lee~\cite{Lopes90}) on intensity values, which were then transformed in amplitude before further processing.

\begin{table}[hbt]
\centering
\caption{Average BDM results for Canny's method, with standard deviation  ​​inside parentheses}\label{aba:tab_BDM_Canny}
\begin{tabular}{l | l | l | l }
\hline
Strategy& No filter   & Torres filter & Enh.\ Lee filter \\ 
\hline
DB-HH   & 26.72 (1.22) & 23.40 (1.92)  & 28.85 (1.86)  \\ 
DB-HV & 30.70 (1.39) & 26.74 (1.33)  & 66.40 (0.42)  \\
DB-VV & 29.81 (2.17) & 26.28 (1.09)  & 24.83 (1.98)  \\ 
ADB & 27.87 (1.16) & 48.16 (0.003)  & 30.24 (1.20)  \\ 
\hline
\end{tabular}
\end{table}

\begin{table}[hbt]
\centering
\caption{Average BDM results for the multi-scale method, with standard deviation  ​​inside parentheses}\label{aba:tab_BDM_multi}
\begin{tabular}{l | l | l | l }
\hline
Strategy& No filter   & Torres filter & Enh.\ Lee filter \\
\hline
DB-HH & 28.23 (0.89) & 28.00 (0.93) & 25.36 (2.11) \\
DB-HV & 28.34 (1.13) & 24.55 (2.97) & 19.56 (2.36) \\ 
DB-VV & 25.62 (1.10) & 28.42 (0.56) & 28.89 (2.25) \\ 
ADB & 25.15 (3.23) & 24.37 (1.52) & 20.71 (3.97) \\ 
\hline
\end{tabular}
\end{table}

\begin{table}[hbt]
\centering
\caption{Average BDM results for the gravitational method; with standard deviation  ​​inside parentheses}\label{aba:tab_BDM_lopez_molina}
\begin{tabular}{l | l | l | l }
\hline
Strategy& No filter & Torres filter & Enh.\ Lee filter \\ 
\hline
DB-HH   & 33.89 (1.98)  & 26.61 (2.00)  &  38.97 (0.79) \\  
DB-HV & 31.95 (0.46)  & 27.14 (1.22)  &  32.26 (2.78) \\ 
DB-VV & 32.35 (1.47)  & 28.95 (0.95)  &  43.65 (1.13) \\ 
DAB & 29.26 (1.62)  & 25.91 (1.55)  &  27.71 (2.62) \\ 
ADB & 31.50 (0.82)  & 26.63 (1.27)  &  18.24 (3.41) \\ 
\hline
\end{tabular}
\end{table}

\begin{table}[hbt]
\centering
\caption{Average BDM results for the gravitational  method modified by Fu's neighbourhood, with standard deviation  ​​inside parentheses}\label{aba:tab_BDM_lopez_molina_Fu}
\begin{tabular}{l | l | l | l }
\hline
Strategy& No filter & Torres filter & Enh.\ Lee filter   \\ 
\hline
DB-HH   & 25.27 (0.76)  & 22.18 (0.48)  & 17.79 (3.05) \\ 
DB-HV & 26.48 (1.00)  & 24.21 (0.65)  & 18.40 (5.75) \\ 
DB-VV & 21.41 (1.97)  & 18.14 (0.77) & 17.83 (2.54) \\ 
DAB & 22.67 (2.29)  & 18.97 (1.62)  &  5.43 (1.68) \\ 
ADB & 23.80 (2.23)  & 22.74 (0.50)  &  5.16 (0.36) \\ 
\hline
\end{tabular}
\end{table}

Tables~\ref{aba:tab_BDM_Canny}, 
\ref{aba:tab_BDM_multi},
\ref{aba:tab_BDM_lopez_molina}, 
and~\ref{aba:tab_BDM_lopez_molina_Fu}
 show the results for BDM mean and standard deviation after applying four methods to twenty  simulated mosaic images: 
 Canny's method,  Lopez-Molina et al's multi-scale method,
 Lopez-Molina et al's original gravitational method, and Lopez-Molina et al's method modified using Fu's $9 \times 9$ neighbourhood.

We see that the best BDM average values were obtained with the use of Lopez-Molina et al's gravitational method modified by Fu's neighbourhood, using the ADB and DAB strategies, both preprocessed with the Enhanced Lee filter. 
Both are significantly higher than the other procedures. 

In Table~\ref{aba:tab_BDM_Canny}, we see that filtering did not have a significative impact on Canny's detector.  
The same is true, to a lesser degree, for most results of the multi-scale and (unmodified) gravitational methods, as can be seen in Tables~\ref{aba:tab_BDM_multi}  and \ref{aba:tab_BDM_lopez_molina}. 
In these methods, there is a slight advantage in preprocessing the images using the Enhanced Lee filter.
However, filtering has an impressive effect on the Lopez-Molina gravitational  method modified with Fu's neighbourhood.
In particular, the best results are obtained for strategies DAB and  ADB with preprocessing with the Enhanced Lee filter. 

\begin{figure}[hbt]
	\centering
a)  \includegraphics[scale=.2,frame]{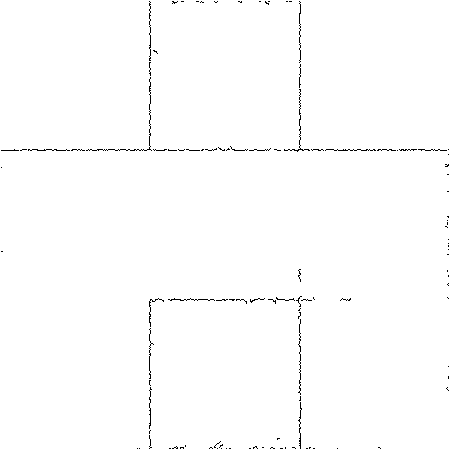}
b) \includegraphics[scale=.2,frame]{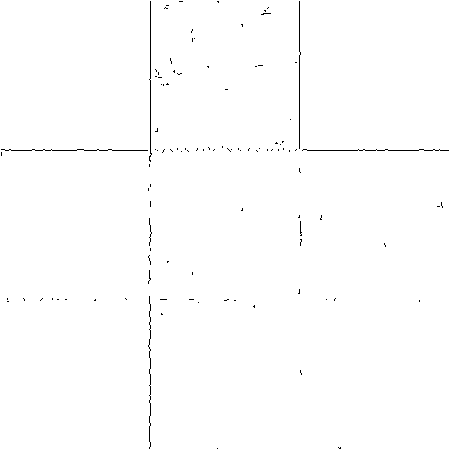} \\
c) \includegraphics[scale=.2,frame]{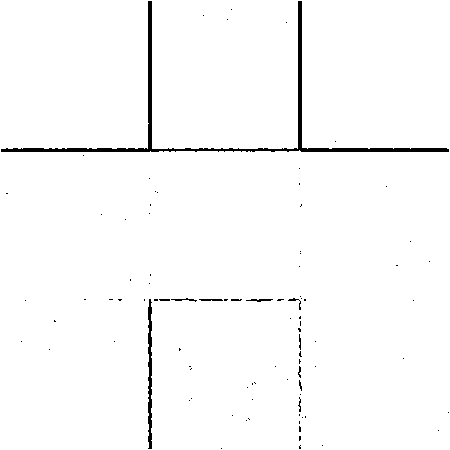}
d) \includegraphics[scale=.2,frame]{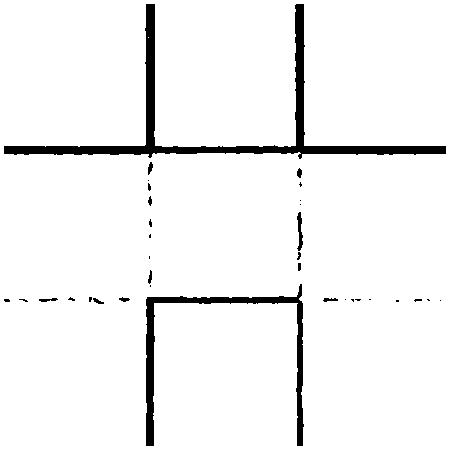} 
	\caption{Best BDM results obtained from the best methods (average): 
	a) Canny (DB-HH, with Torres filtering, BDM = 18.51), 
	b) Multi-scale (DB-HV, with Enh.\ Lee filtering, BDM = 14.94), 
	c) Gravitational (ADB with Enh.\ Lee filtering, BDM = 10.96) and 
	d) Gravitational and Fu (ADB with Enh.\ Lee filtering, BDM = 3.05) 
	}
	\label{fig:best-meth}
\end{figure}

Figure~\ref{fig:best-meth}  shows 
 the negative images corresponding to the best results, according to BDM, obtained by the edge detection methods and the filtering strategies with the best average values; note that the image boundaries are depicted only for illustrative purposes. 
We see that according to BDM the best binary image (depicted in Figure~\ref{fig:best-meth}b) presents little noise and most of the regions are separated, even though the lines are rather thick. We also see that BDM was able to distinguish the best image from the others. 

Figure \ref{fig:best-meth}b shows the best results from the methods come from filtered images, which raises the question of how important preprocessing  by filtering is. 
In what follows we discuss the details of  the  gravitational method using the original $3 \times 3$ and  Fu's $9 \times 9$ neighbourhood in relation to filtering. We take the simulation that obtained the best BDM results for each type of neighbourhood. We see in these examples, that filtering does, indeed, ameliorate the results for all methods.

\begin{figure}[hbt]
	\centering
a)   \includegraphics[scale=.2,frame]{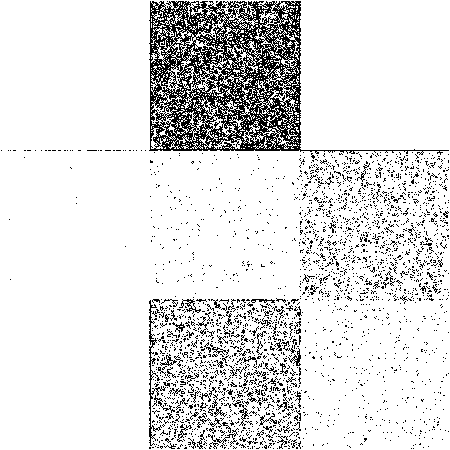}
b)   \includegraphics[scale=.2,frame]{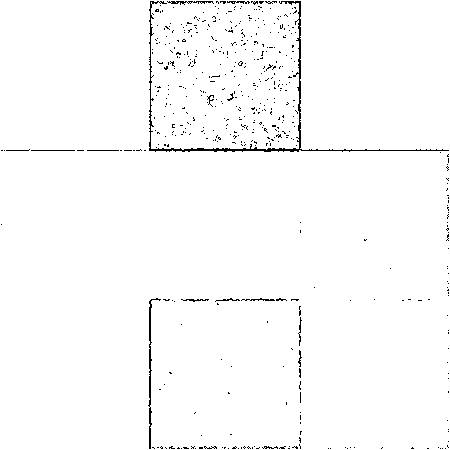}
c)   \includegraphics[scale=.2,frame]{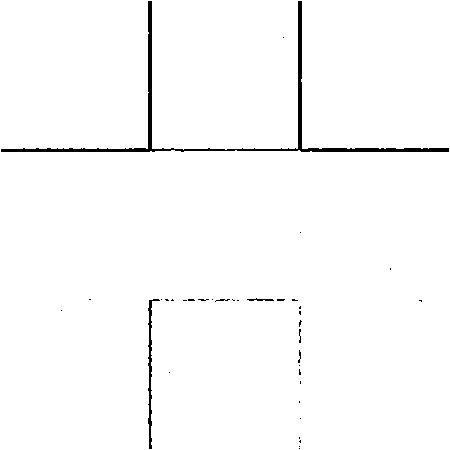}
	\caption{Results for the gravitational method with original 3$\times$3 window, on a single simulated image from ADB: 	 
	a) no filtering (BDM = 31.90), b) Torres (BDM = 27.73) and c) Enh.\ Lee (BDM = 10.96)
	}
	\label{fig:test-bustince}
\end{figure}

\begin{figure}[hbt]
	\centering
a)   \includegraphics[scale=.2,frame]{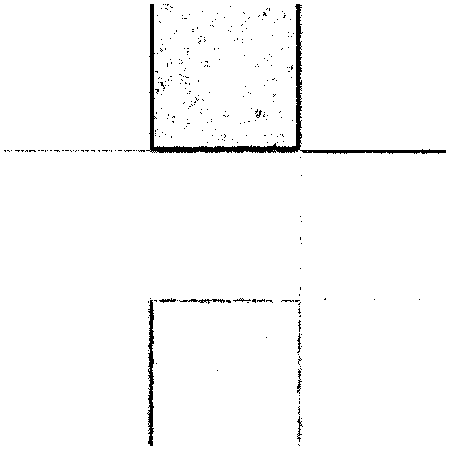}
b)   \includegraphics[scale=.2,frame]{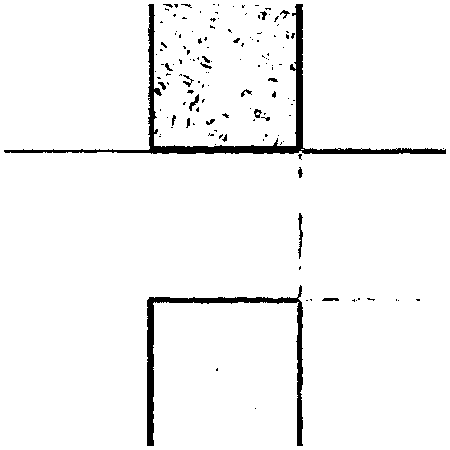}
c)   \includegraphics[scale=.2,frame]{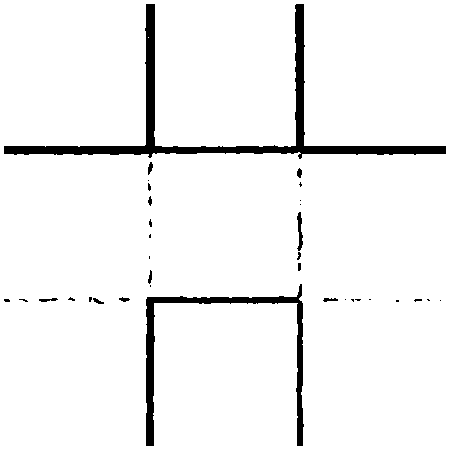}
	\caption{Results for the gravitational method modified with Fu's neighbourhood, on a single simulated image from ADB:  
	a) no filtering (BDM = 23.80), b) Torres (BDM = 23.76) and c) Enh.\ Lee (BDM = 3.05)
	}
	\label{fig:test-bustincefu}
\end{figure}

Figures~\ref{fig:test-bustince}.a), \ref{fig:test-bustince}.b) and \ref{fig:test-bustince}.c) respectively show that: using the $3 \times 3$ neighbourhood for the original gravitational method, the unfiltered image is very noisy; the Torres filter reduced the noise and separated the regions; and the Lee filter detected false edges. 
In Figure \ref{fig:test-bustincefu},  we see that Fu's $9 \times 9$ neighbourhood  detected almost all the edges, especially when using Lee's filter. 
Filtering for the modified method presented a larger trade-off between detection of edges and reduction of noise  (some edges were detected using Torres filter  with an increase of noise when compared to the unfiltered image).

When  we compare the results in Figures~\ref{fig:test-bustince} and~\ref{fig:test-bustincefu} we see that the gravitational method modified with Fu's $3\times3$ neighbourhood clearly produced better results than the method with the original $3\times3$ window, which agrees with the BDM evaluation. 

\begin{figure}[hbt]
	\centering
\includegraphics[scale=.4,frame]{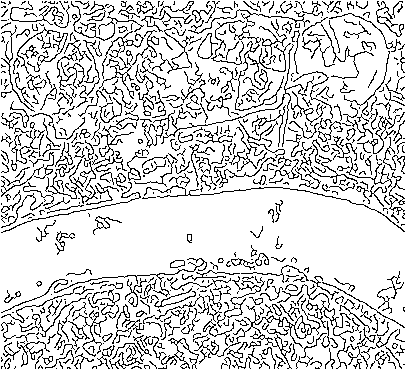}
\includegraphics[scale=.4,frame]{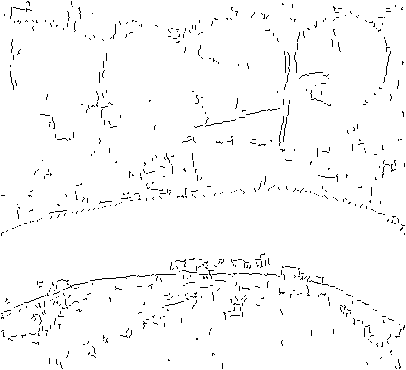}
\includegraphics[scale=.4,frame]{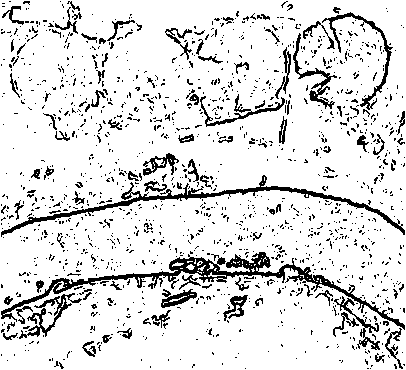}
\includegraphics[scale=.4,frame]{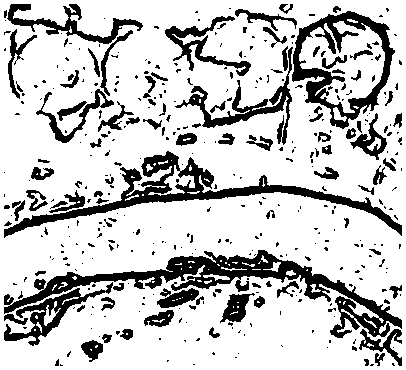}
\caption{ HV Bebedouro binary images, with Enh.\ Lee filter:
the first row depicts results obtained using the Canny and the Multi-scale methods; 
and the second row depicts results obtained using the original Gravitational method and its modification with Fu's neighbourhood, 
(the latter  two methods use binarization threshold=.2)
	}
	\label{fig:bebedouro-hv}
\end{figure}

\begin{figure}[hbt]
	\centering
  \includegraphics[scale=.4,frame]{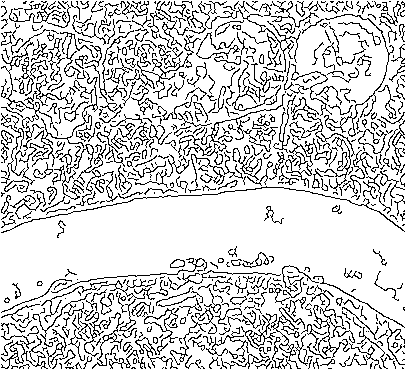}
  \includegraphics[scale=.4,frame]{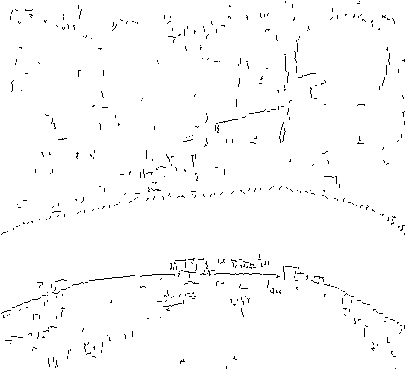}
  \includegraphics[scale=.4,frame]{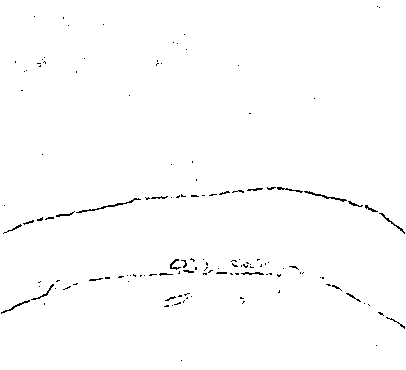}
  \includegraphics[scale=.4,frame]{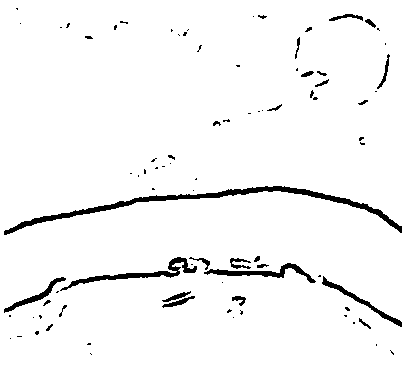}
\caption{ADB Bebedouro binary images, with Enh.\ Lee filter:
the first row depicts results obtained using the Canny and the Multi-scale methods; 
and the second row depicts results obtained using the original Gravitational method and its modification with Fu's neighbourhood, 
(the latter  two methods use binarization threshold=.2)
	}
	\label{fig:bebedouro-adb}
\end{figure}

\begin{figure}[hbt]
	\centering
  \includegraphics[scale=.4,frame]{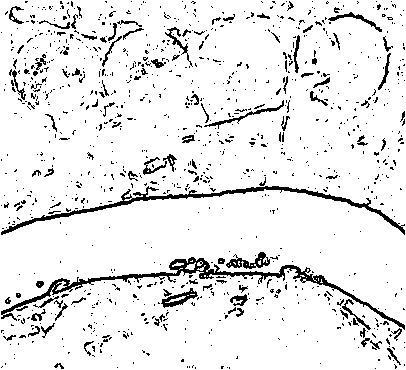}
  \includegraphics[scale=.4,frame]{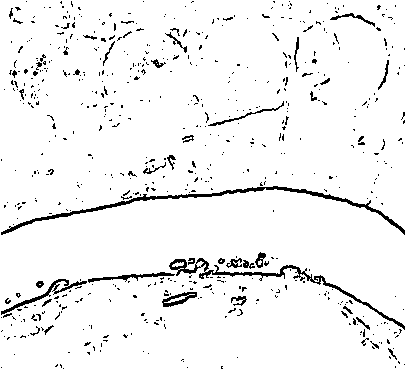}
  \includegraphics[scale=.4,frame]{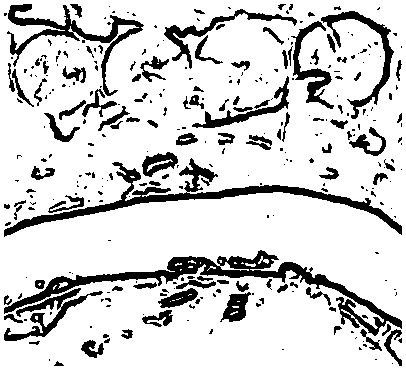}
  \includegraphics[scale=.4,frame]{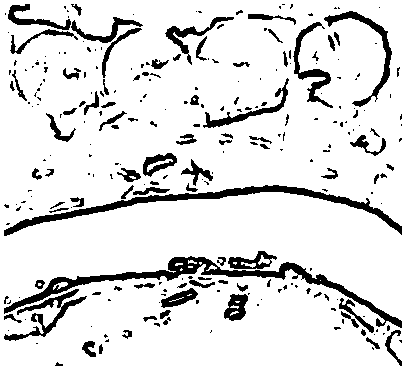}
\caption{ DAB Bebedouro binary images, with Enh.\ Lee filter: 
the first and second rows respectively depict results using the original Gravitational and Gravitational modified with Fu's neighbourhood, 
the first and second columns respectively depict results obtained with binarization thresholds .1 and .2
	}
	\label{fig:bebedouro-dab}
\end{figure}

Figures \ref{fig:bebedouro-hv}, \ref{fig:bebedouro-adb}, and \ref{fig:bebedouro-dab}  show  the best edge detection results obtained for the Bebedouro SAR image  in terms of  visual analysis, with the application of the Enhanced Lee filter, for the parameterizations used here. 
The figures  present the results for HV polarization; in general, HH (respec.\ VV) polarization produced images with more noise (respec.\ less information) than HV.
Figures~\ref{fig:bebedouro-adb} and~\ref{fig:bebedouro-dab}, 
respectively, show the results of the application of  ADB and DAB aggregation strategies.  
ADB in general produced binary images with very little information for all methods; the decrease of noise in relation to the individual polarizations does not compensate the lack of information.
In general, the DBA aggregation method produced results with less noise for the gravitational method, with and without modification, than the results obtained with the individual polarizations. The best results for the gravitational method, 
both with and without modification with Fu's neighbourhood, were obtained with thresholds around the same interval that produced the best results using the mosaics.

\section{Conclusions}\label{sec:conclusions}

Contrary to what happens with optical imagery, few algorithms are specifically dedicated to PolSAR image edge detection~\cite{Fu2012}.
One interesting means to create edge detection algorithms for SAR images is to modify those created for optical images, in such a way as to reduce the non-Gaussian noise. 
Here we have investigated the modification of a method issued from Computational Intelligence for optical imagery,  the  gravitational edge detection method extension proposed in \cite{LopezMolina2010} (see also \cite{Sun2007}), to synthetic aperture radar imagery. 
In order to deal with speckle, we modified the gravitational method with a non-standard $9 \times 9$ neighbourhood configuration proposed by Fu et al~\cite{Fu2012}: considering a $3 \times 3$ window centered around a given pixel,  the value of any pixel in the window becomes the average value of the region associated to that  pixel in the non-standard neighbourhood configuration. 

Considering that SAR imagery has different polarizations, and that their joint use may compensate for the presence of speckle, we also proposed a typology of experiments regarding aggregation of these images.  
In particular, we addressed two procedures:  DAB (edge Detection on non-binary images, Aggregation of the resulting non-binary  images, Binarization) and
ADB (Aggregation of non-binary  images, edge Detection on the resulting non-binary  image, Binarization). 

For means of comparison, we also addressed the use of two other edge detector methods  
stemming from the realm of optical images: the traditional method proposed by Canny~\cite{CANNY1986} 
and a recent multi-scale one coming from Computational Intelligence, based on 
 Sobel operators for edge extraction and the concept of Gaussian scale-space  \cite{lopez-molina-13-multiscale}. 
 
We studied the effect of filtering the images prior to edge detection by two procedures: 
Enhanced Lee~\cite{Lopes90} and Torres~\cite{TorresPolarimetricFilterPatternRecognition} filters.
The methods were applied on twenty samples of a scene, which were simulated using Wishart distributions derived from a fully polarimetric image~\cite{Barrreto2013}.
Using both visual inspection and the Baddeley Delta metric~\cite{Baddeley1992}
we verified that the combination with the Lopez-Molina technique with the $9 \times 9$ neighbourhood proposed by Fu et al~\cite{Fu2012} and preprocessing with the Enhanced Lee filter produced the best results. 

This paper is an extended version of \cite{SilvaFLINS14}; together, these studies represent a first step towards investigating the use of edge detection methods derived from Computational Intelligence techniques for use in SAR images. 
The main implication of our  results is that the joint use filtering and neighbourhood modification on those methods, as well as the use of aggregation of the individual polarization images, are able to deal with speckle, which is crucial when detecting edges in radar imagery.

Future work includes modifying the Lopez-Molina method with other types of neighbourhoods, such as Nagao-Matsuyama~\cite{Nagao1979}; to verify the performance of other T-norms than  the product to calculate the gravitational forces; and to perform preprocessing with other filters. 
We also intend to investigate the use of the proposed procedure  with other edge detection methods, such as the one described in~\cite{LopezMolina2011}, involving fuzzy sets. 

We would like to better address the issue of aggregation. 
Here we have dealt exclusively with the aggregation of non-binary images, using the arithmetic means in strategies DAB and ADB. 
In the future, we intend to explore aggregation of the images considering families of operators in general, such as weighted means, ordered weighted means (OWA), T-norms, and T-conorms~\cite{DP88}. 
Also, we intend to  study other operators than the average to perform aggregation of  pixel values in regions of a non-standard neighbourhood. 

Moreover, we intend to assess the results using other methods than BDM,  such as the one proposed recently by Frery et al~\cite{Buemi2014}. 
We would also like to draw comparisons with other edge detection algorithms, such as the one proposed by Fu in 2012~\cite{Fu2012}.  
 
Last but not least, we intend to verify the use of the approach considering fully polarimetric images (PolSAR), instead of just intensity images. 
In this case, Torres filter, designed specifically  for PolSAR images, can be more adequately employed.

\section*{Acknowledgements}
\small
The authors are  grateful for 
Wagner Barreto Silva, Leonardo Torres and Corina Freitas for help in the preparation of this manuscript. 
They are also thankful for the editor and reviewers for comments and suggestions.
The Brazilian authors acknowledge support from CNPq (Projeto Universal 487032/2012-8).
The Spanish authors have been supported by project TIN2013-40765-P of the Spanish Government.

\section*{References}

\bibliography{posFlins14Pedro}

\end{document}